\documentclass[journal]{IEEEtran} 
\usepackage{amssymb}
\usepackage{graphicx}
\usepackage{textcomp}
\usepackage{amsmath}

\usepackage{color}
\usepackage{soul}
\usepackage{url}
\usepackage{cite}

\usepackage{xcolor}
\usepackage{framed}

\usepackage{graphicx}
\usepackage{graphics}
\usepackage{multirow}
\usepackage{caption}
\usepackage{subcaption}
\usepackage{gensymb}

\usepackage{multirow}
\usepackage{array}
\usepackage{booktabs}
\usepackage{tabularx}
\newcolumntype{L}[1]{>{\raggedright\let\newline\\\arraybackslash\hspace{0pt}}m{#1}}
\newcolumntype{C}[1]{>{\centering\let\newline\\\arraybackslash\hspace{0pt}}m{#1}}
\newcolumntype{R}[1]{>{\raggedleft\let\newline\\\arraybackslash\hspace{0pt}}m{#1}}

\usepackage{blindtext}

\newcommand{\comst}[1]{}

\newcommand{\com}[1] {}
\newcommand{\mytilde}{\raise.17ex\hbox{$\scriptstyle\mathtt{\sim}$}}

\colorlet{shadecolor}{blue!20}

\begin{document}

\title{EndoNet: A Deep Architecture for Recognition Tasks on Laparoscopic Videos}

\author{Andru P. Twinanda, Sherif Shehata, Didier Mutter, Jacques Marescaux, Michel de Mathelin,  Nicolas Padoy

\thanks{Andru P. Twinanda, Sherif Shehata, Michel de Mathelin, and Nicolas Padoy are affiliated with ICube, University of Strasbourg, CNRS, IHU Strasbourg, France (email: twinanda@unistra.fr)}
\thanks{Didier Mutter and Jacques Marescaux are affiliated with the University Hospital of Strasbourg, IRCAD and IHU Strasbourg, France.}
}%

\maketitle

\begin{abstract}

Surgical workflow recognition has numerous potential medical applications, such as the automatic indexing of surgical video databases and the optimization of real-time operating room scheduling, among others. As a result, phase recognition has been studied in the context of several kinds of surgeries, such as cataract, neurological, and laparoscopic surgeries. In the literature, two types of features are typically used to perform this task: visual features and tool usage signals. However, the visual features used are mostly
handcrafted. Furthermore, the tool usage signals are usually collected via a manual annotation process or by using additional equipment.
In this paper, we propose a novel method for phase recognition that uses a convolutional neural network (CNN) to automatically learn features from cholecystectomy videos and that relies uniquely on visual information.  
In previous studies, it has been shown that the tool usage signals can
provide valuable information in performing the phase recognition task. 
Thus, we present a novel CNN architecture, called EndoNet, that
is designed to carry out the phase recognition and tool presence detection
tasks in a multi-task manner. 
To the best of our knowledge, this is the first work proposing to use a CNN for multiple recognition tasks on laparoscopic
videos. Extensive experimental comparisons to other methods show that  EndoNet yields state-of-the-art results for both tasks.

\end{abstract}

\section{Introduction}

\IEEEPARstart{I}{}n the community of computer-assisted interventions (CAI), recognition
of the surgical workflow is an important topic because it offers solutions
to numerous demands of the modern operating room (OR) \cite{cleary_OR2020}. For instance, such recognition is an essential component to develop context-aware systems  that can monitor the surgical processes, optimize OR and staff scheduling, and provide automated assistance to the clinical staff.  With the ability to segment surgical workflows, it would also be possible to automate the indexing of surgical video databases, which is currently a time-consuming manual process. In the long run, through finer analysis of the video content, such context-aware systems could also be used to alert the clinicians to probable upcoming complications.

Various types of features have been used in the literature to carry
out the phase recognition task. For instance, in \cite{bouarfa_jbmi2011,padoy_mia2012},
binary tool usage signals were used to perform phase recognition on
cholecystectomy procedures. In more recent studies \cite{katic_ipcai2014,forestier_bmi2013},
surgical triplets (consisting of the utilized tool, the anatomical
structure, and the surgical action) were used to represent the frame
at each time step in a surgery. However, these features are typically
obtained through a manual annotation process, which is virtually
impossible to perform at test time. Despite existing efforts \cite{lalys_ijcars2012}, it is still an open question whether such information can be obtained reliably in an automatic manner.

Another feature type that is typically used to perform the phase
recognition task is visual features, such as pixel values and
intensity gradients \cite{blum_miccai2010}, spatio-temporal features
\cite{zappella_mia2013}, and a combination of features (color, texture,
and shape) \cite{lalys_tbme2012}. However, these features are handcrafted,
i.e., they are \textit{empirically} designed to capture certain information
from the images, leading to the loss of other possibly significant characteristics
during the feature extraction process. 

In this paper, we present a novel method for phase recognition that overcomes the afore-mentioned limitations.\\
\textbf{First}, instead of
using handcrafted features, we propose to learn inherent visual features from surgical (specifically cholecystectomy) videos
to perform phase recognition. We focus on visual features because videos
are typically the only source of information that is readily available
in the OR. In particular, we propose to learn the features using a convolutional neural
network (CNN), because CNNs have dramatically improved the results for various image recognition tasks in recent years, such
as image classification \cite{krizhevsky_nips2012} and object detection
\cite{girshick_cvpr2014}. In addition, it is advantageous to automatically
learn the features from laparoscopic videos because of the visual
challenges inherent in them, which make it difficult to design suitable features.
For example, the camera in laparoscopic procedures
is not static, resulting in motion blur and high variability of
the observed scenes along the surgery. The lens is also often stained
by blood which can blur or completely occlude  the scene captured by
the laparoscopic camera. \\
\textbf{Second}, based on our and others' promising results of
 using tool usage signals to perform phase recognition \cite{stauder_ipcai2014,padoy_mia2012},
we hypothesize that tool information can be additionally utilized to generate
more discriminative features for the phase recognition task. This has also been shown in \cite{blum_miccai2010}, where the tool usage signals are used to reduce the dimension of the handcrafted visual features through canonical correlation analysis (CCA) in order to obtain more semantically meaningful and discriminative features. To incorporate the tool information,
we propose to implement a multi-task framework in the feature learning process. 
The resulting CNN architecture, that we call EndoNet, is designed to jointly perform  
the phase recognition and tool presence detection
tasks. The latter is the task of automatically determining all types
of  tools present in an image. 
In addition to helping EndoNet 
learn more discriminative features, the tool presence detection task itself is also interesting to perform because it could 
be exploited for many applications, for instance to automatically index a surgical video database by  
labeling the tool presence in the videos. Combined with other signals, it could also 
be used to identify a potential upcoming complication 
by detecting tools that should not appear in a certain phase. 
It is important to note that this
task differs from the usual tool detection task \cite{sznitman_miccai2014}, 
because it does not require tool localization. In
addition, the tool presence is solely determined by the visual information
from the laparoscopic videos. Thus, it does not result in the same tool information as the one
used in \cite{padoy_mia2012}, which cannot always be obtained
from the laparoscopic videos alone. For example, the presence of trocars used in \cite{padoy_mia2012}
is not always apparent in the laparoscopic videos. Automatic presence detection for such tools would require another source of information, e.g., an external video. 

Training CNN architectures requires a substantial capacity of parallel
computing and a large amount of labeled data. In the domain of medicine, labeled 
data is particularly difficult to obtain due to regulatory restrictions and 
the cost of manual annotation. Girshick et al. \cite{girshick_cvpr2014} recently 
showed that transfer learning can be 
used to train a network when labeled data is scarce. Inspired by \cite{girshick_cvpr2014}, we 
perform transfer learning to train the proposed EndoNet architecture.

To validate our method, we build a large dataset of cholecystectomy
videos containing 80 videos recorded at the University Hospital of Strasbourg. In addition, to demonstrate
that our proposed (i.e., EndoNet) features are generalizable, we carry out additional experiments 
on the EndoVis workflow challenge dataset%
\footnote{http://grand-challenge.org/site/endovissub-workflow/data/%
} containing seven cholecystectomy videos recorded at the Hospital Klinikum Rechts der Isar in Munich. Through extensive comparisons, we also show that EndoNet outperforms other state-of-the-art methods. Moreover, we also demonstrate
that training the network in a multi-task manner results in a better network than training in a single-task manner.

In summary, the contributions of this paper are five-fold: (1)
for the first time, CNNs are utilized to extract visual features for
recognition tasks on laparoscopic videos, (2) we design a CNN architecture
that jointly performs the phase recognition and tool presence detection
tasks, (3) we present a wide range of comparisons between our method and other approaches, (4) we show state-of-the-art results for
both tasks on cholecystectomy videos using solely visual features, and (5) we demonstrate the feasibility of using EndoNet in addressing several practical CAI applications.

\section{Related Work}

\subsection{Tool Presence Detection\label{sub:Tool-Presence-Detection}}

The literature addressing the problem of automatic tool \textit{presence}
detection in the CAI community is
still limited. The approaches typically
focus on other tasks, such as tool detection \cite{sznitman_miccai2014,bouget_tmi2015},
tool pose estimation \cite{allan_miccai2015}, and tool tracking 
\cite{rieke_miccai2015,reiter_miccai2012}. In addition, most of the methods are only 
tested on short sequences, while we carry out the task on the complete procedures.

In recent studies \cite{kranzfelder_sr2013,neumuth_ijcars2012}, radio frequency
identification (RFID)-tagged surgical tools have been proposed for
tool detection and tracking. Such an active tracking system can be
used to solve the tool presence detection problem, but this system
is complex to integrate into the OR. Thus,
it is interesting to investigate other features that are already available
in the OR, e.g., visual cues from the videos. For instance, in \cite{speidel_spie2009}, Speidel et al. presented an approach
to automatically recognize the types of the tools that appear in laparoscopic
images. However, the method consists of many steps, such
as tool segmentation and contour processing. In addition, it also
requires the 3D models of the tools to perform the tool categorization.
In a more recent work \cite{lalys_tbme2012}, Lalys et al. proposed
to use an approach based on the Viola-Jones object detection framework
to automatically detect the tools in cataract surgeries, such as the knife
and Intra Ocular Lens instruments. However, the tool presence detection
problem on laparoscopic videos poses other challenges that do not
appear in cataract surgeries where the camera is static and the tools
are not articulated. In this paper, we propose
a more direct approach to perform the tool presence detection task by
using only visual features without localization steps.

\subsection{Phase Recognition}

The phase recognition task has been addressed in several types of
surgeries, ranging from cataract \cite{lalys_tbme2012,quellec_tmi2015},
neurological \cite{forestier_bmi2013}, to laparoscopic
surgeries \cite{katic_ipcai2014,blum_miccai2010,lo_miccai2003}.
Multiple types of features have also been explored to carry out the
task, such as tool usage signals \cite{padoy_mia2012,forestier_bmi2013},
surgical action triplets \cite{forestier_ijcars2015,katic_ipcai2014},
and visual features \cite{blum_miccai2010,lea_amia2013}. Since
we propose to carry out the task relying solely on the visual features,
we focus the literature discussion on methods that use the visual features. 

In \cite{padoy_iaai2008}, Padoy et al. proposed an online
phase recognition method based on Hidden Markov Model (HMM) 
that combines the tool usage signals and two visual
cues from the laparoscopic images. The first and second cues respectively
indicate whether the camera is inside the patient's body and whether
clips are in the field of view. However, to recognize the phase, this
method requires the tool signals which are not always immediately available
in the OR. Instead, Blum et al. \cite{blum_miccai2010} proposed to use the tool usage signals to perform dimensionality reduction on the visual features using CCA. 
Once the projection function is obtained, the tool information is not
required anymore to estimate the surgical phase. At test time, the
visual features are mapped to the common space and then later used
to determine the phase. The method performed well, resulting in an accuracy of 76\%. However, 
it has only been tested on a dataset of 10 videos. In addition,
the method is potentially limited by the choice of handcrafted features
that are used: horizontal and vertical gradient magnitudes, histograms
and the pixel values of the downsampled image. 

In a more recent work \cite{lalys_tbme2012}, Lalys et al. presented
a framework to recognize high-level surgical tasks for cataract surgeries
using a combination of visual information: shape, color, texture,
and mixed information. The features also contain the tool presence
information which is automatically extracted from the microscopic videos,
as mentioned in Subsection \ref{sub:Tool-Presence-Detection}. By
using HMM on top of the features, the method yields 91\% accuracy.
However, the method was evaluated on cataract surgeries, which are
substantially different from cholecystectomy surgeries. Cholecystectomy
surgeries are generally longer than cataract surgeries. In addition,
cholecystectomy videos have visual challenges that are not present
in cataract surgeries, such as rapid camera motions, the presence of smoke, and
the presence of more articulated tools. In \cite{lea_wacv2015}, Lea et al. used 
skip-chain conditional random field on top of kinematic and image features 
to segment and recognize fine-grained surgical activities, such as needle 
insertion and tying knot. However, the method is tested on a dataset that 
contains short sequences (around two minutes). Furthermore, the visual
features that are utilized in the afore-mentioned methods are handcrafted. 

In \cite{klank_ijcars2008}, Klank et al. proposed to learn automatically the visual
features from cholecystectomy videos to carry out the
phase recognition task. The approach is based on genetic programming
that mutates and crosses the features using predefined operators. The method is therefore limited by the set of predefined
operators. In addition, the learnt features failed to give better
recognition results than the handcrafted features in some cases.

\subsection{Convolutional Neural Networks}

In the computer vision community, convolutional neural networks (CNNs)
are currently one of the most successful feature learning methods in
performing various tasks. For instance, 
Krizhevsky et al. \cite{krizhevsky_nips2012} addressed the image classification problem on the massive
ImageNet database \cite{imagenet} by proposing to use a CNN architecture, referred to as AlexNet. 
They showed that the features
learnt by the CNN dramatically improve the classification results
compared to the state-of-the-art handcrafted features, e.g., Fisher
Vector on SIFT \cite{sanchez_cvpr2011}. Furthermore, in \cite{donahue_corr2013}, it has been shown 
that the network trained in \cite{krizhevsky_nips2012} is so powerful that 
it can be used as a black-box feature extractor 
(without any modification) to successfully perform several tasks, 
including scene classification and domain adaptation.  

CNNs are hard to train because they typically contain a high number of unknowns. For instance, the AlexNet architecture contains over 60M parameters. It is essential to have a high computational power and a huge amount of annotated data to train the networks. Recently, Girshick et al. \cite{girshick_cvpr2014} showed that a new network can be learnt despite the scarcity of labeled data by performing transfer learning. 
They proposed to take a pre-trained CNN model as initialization and fine-tune 
the model to obtain a new network. It is shown that the fine-tuned network yielded a state-of-the-art performance for object recognition task, despite being fine-tuned on a network trained
for image classification.

\section{Methodology}

\begin{figure}
\begin{centering}
\includegraphics[width=8cm]{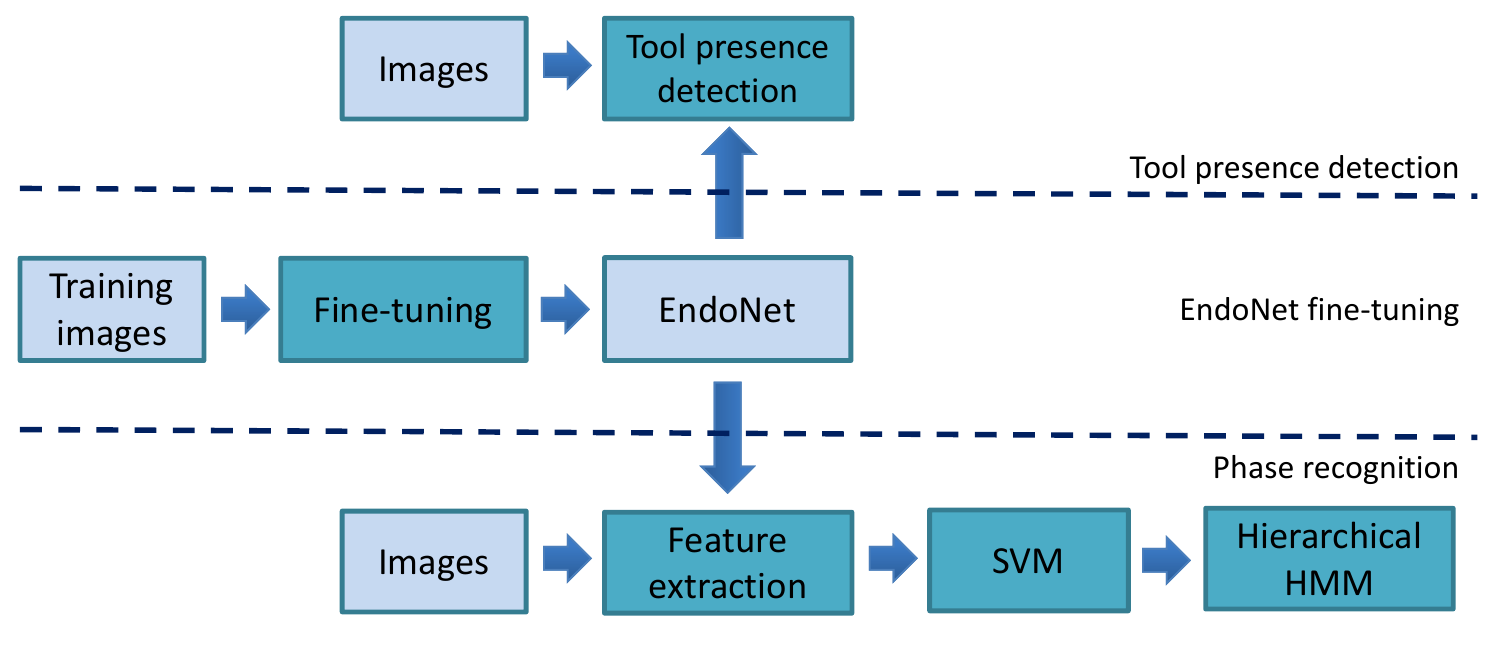}
\par\end{centering}

\caption{Full pipeline of the proposed approach. \label{fig:Phase-recognition-pipeline.} }
\end{figure}

The complete pipeline of our proposed approach is shown in Fig. \ref{fig:Phase-recognition-pipeline.}.
The first step is to train the EndoNet architecture via a fine-tuning process.
Once the network is trained, it is used for both the tool presence detection and phase recognition tasks. For the former, the confidence given by the network is directly used to perform the task. For the latter, the network is used to extract the visual features from
the images. These features are then passed to the Support Vector
Machine (SVM) and Hierarchical HMM to obtain
the final estimated phase. 

\subsection{EndoNet Architecture}

\begin{figure*}
\begin{centering}
\includegraphics[width=18cm]{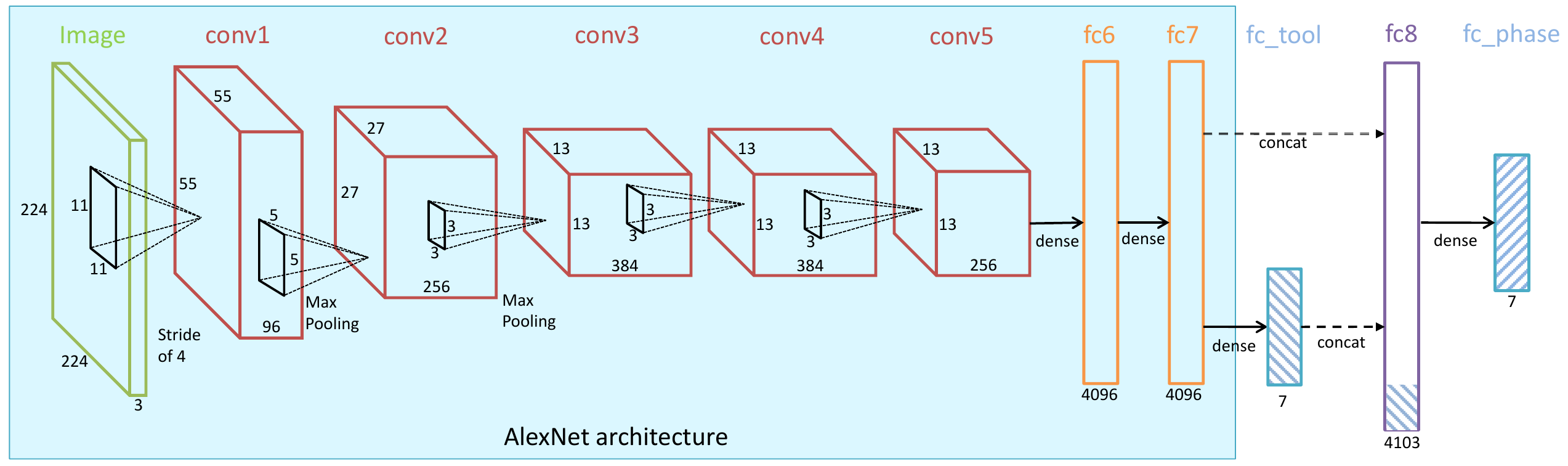}
\par\end{centering}

\caption{EndoNet architecture (best seen in color). The layers shown in the turquoise rectangle are the same as in the AlexNet architecture.
 \label{fig:CammaNet-Multi-architecture.}}
\end{figure*}

The EndoNet architecture is designed based on two assumptions, which will be confirmed by the experiments presented in Section \ref{sec:Exps}:
\begin{itemize}
\item more discriminative features for the phase recognition task can be learnt
from the dataset if the network is fine-tuned in a multi-task manner,
i.e., if the network is optimized to carry out not only phase recognition,
but also tool presence detection;
\item since the tool signals have been successfully used to carry out
phase recognition in previous work \cite{padoy_mia2012,forestier_bmi2013,lalys_tbme2012},
the inclusion of automatically generated tool detection signals in the final feature can improve the recognition.
\end{itemize}
The proposed EndoNet architecture is shown in Fig. \ref{fig:CammaNet-Multi-architecture.}.
The architecture is an extension of the AlexNet architecture \cite{krizhevsky_nips2012},
which consists of an input layer (in green), five convolutional layers
(in red, $\mathtt{\mathtt{conv}1}$-$\mathtt{conv5}$), and two fully-connected 
layers (in orange, $\mathtt{fc6}$-$\mathtt{fc7}$). The
output of layer $\mathtt{fc7}$ is connected to a fully-connected
layer $\mathtt{fc}$\_$\mathtt{tool}$, which performs the tool presence
detection. Since there are seven tools defined in the dataset used to train the network, 
the layer $\mathtt{fc}$\_$\mathtt{tool}$ contains 7
nodes, where each node represents the confidence for a tool to be present
in the image. This confidence is later concatenated with the output
of layer $\mathtt{fc7}$ in layer $\mathtt{fc8}$ to construct the
final feature for the phase recognition. Ultimately, the output of
layer $\mathtt{fc8}$ is connected to layer $\mathtt{fc}$\_$\mathtt{phase}$
containing 7 nodes, where each node represents the confidence that an
image belongs to the corresponding phase. The surgical tool types and the surgical
phases are described in Subsection \ref{sub:Dataset}.

\subsection{Fine-Tuning}

The network is trained using stochastic gradient descent with
two loss functions defined for the tasks. The tool presence
detection task is formulated as $N_{t}$ binary classification tasks,
where $N_{t}=7$ is the number of tools. For each binary classification
task, the cross-entropy function is used to compute the loss. Thus
for $N_{i}$ images in the batch, the complete loss function of the tool presence
detection task for all tools is defined as:

\begin{equation}
\mathcal{L_{T}}=\frac{-1}{N_{i}}\sum_{t=1}^{N_{t}}\sum_{i=1}^{N_{i}}\left[k_{t}^{i}\log\left(\sigma\left(v_{t}^{i}\right)\right)+\left(1-k_{t}^{i}\right)\log\left(1-\sigma\left(v_{t}^{i}\right)\right)\right],\label{eq:ToolLoss}
\end{equation}
where $i\in\left\{ 1,\ldots,N_{i}\right\} $ and $t\in\left\{ 1,\ldots,N_{t}\right\} $
are respectively the image and tool indices, $k_{t}^{i}\in\{0,1\}$
and $v_{t}^{i}$ are respectively the ground truth of tool presence
and the output of layer $\mathtt{fc}$\_$\mathtt{tool}$ corresponding
to tool $t$ and image $i$, and $\sigma\left(\cdot\right)\in(0,1)$
is the sigmoid function.

Phase recognition is regarded as a multi-class
classification task. The softmax multinomial logistic function, which
is an extension of the cross-entropy function, is utilized to compute
the loss. The function is formulated as:

\begin{equation}
\mathcal{L_{P}}=\frac{-1}{N_{i}}\sum_{i=1}^{N_{i}}\sum_{p=1}^{N_{p}}l_{p}^{i}\log\left(\varphi\left(w_{p}^{i}\right)\right),\label{eq:PhaseLoss}
\end{equation}
where $p\in\left\{ 1,\ldots,N_{p}\right\} $ is the phase index and $N_{p}=7$ is the number of phases, $l_{p}^{i}\in\{0,1\}$
and $w_{p}^{i}$ are respectively the ground truth of the phases
and the output of layer $\mathtt{fc}$\_$\mathtt{phase}$ corresponding
to phase $p$ and image $i$, and $\varphi\left(\cdot\right)\in[0,1]$
is the softmax function.

The final loss function is the summation of both losses: $\mathcal{L}=a\cdot\mathcal{L_{T}}+b\cdot\mathcal{L_{P}}$, 
where $a$ and $b$ are weighting coefficients. In this work, we set $a=b=1$ as preliminary experiments have shown no improvement when varying these parameters.
One should note that assigning either $a=0$ or $b=0$ is equivalent to designing a CNN that is optimized to carry out only the phase recognition task or the tool presence detection task, respectively.

\subsection{SVM and Hierarchical HMM}

The output of layer $\mathtt{fc8}$ is taken as the image feature.
These features are used to compute confidence values $\mathbf{v}_{p}\in\mathbb{R}^{7}$ for phase estimation
using a one-vs-all multi-class SVM. Since the confidence $\mathbf{v}_{p}$ 
is obtained without taking into account any temporal information, 
it is necessary to enforce the temporal constraint of the
surgical workflow. Here, we use use an extension of HMM, namely
a two-level Hierarchical HMM (HHMM) \cite{padoy_cvw2009}. The top-level 
contains nodes that model the inter-phase dependencies, while
the bottom-level nodes model the intra-phase dependencies. We train the
HHMM adopting the learning process presented in \cite{padoy_cvw2009}.
Here, the observations are given by the confidence $\mathbf{v}_{p}$
from the SVM. For offline recognition, the Viterbi algorithm \cite{viterbi}
is used to find the most likely path through the HHMM states. As
for online recognition, the phase prediction is computed using the
forward algorithm.


One can observe that EndoNet already provides confidence values through the output of layer $\mathtt{fc}$\_$\mathtt{phase}$, thus it is not essential to pass EndoNet features to the SVM to obtain the confidence values $\mathbf{v}_{p}$. Furthermore, in preliminary experiments, we observed that there was only a slight difference of performance between $\mathbf{v}_{p}$ and $\mathtt{fc}$\_$\mathtt{phase}$ in recognizing the phases both before and after applying the HHMM.  However, this additional step is necessary in order to provide a fair comparison with other features, which are passed to the SVM to obtain the confidence. In addition, using the output of layer $\mathtt{fc}$\_$\mathtt{phase}$ as the phase estimation confidence is only applicable to datasets that share the same phase definition as the one in the fine-tuning dataset.
Thus, this step is also required for the evaluation of the network generalizability to other datasets that might have a different phase definition.

\section{Experimental Setup}

\subsection{Dataset \label{sub:Dataset}}

\begin{figure*}
\begin{centering}
\begin{tabular}{c}
\includegraphics[width=2.5cm,height=1.9cm]{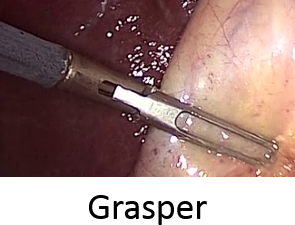} \includegraphics[width=2.5cm,height=1.9cm]{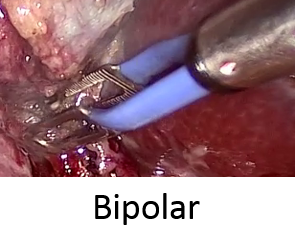}
\includegraphics[width=2.5cm,height=1.9cm]{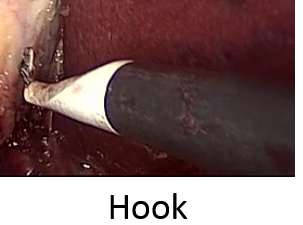} \includegraphics[width=2.5cm,height=1.9cm]{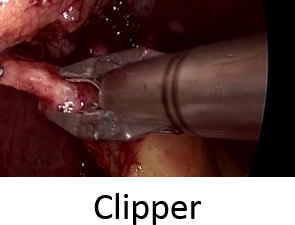}
\includegraphics[width=2.5cm,height=1.9cm]{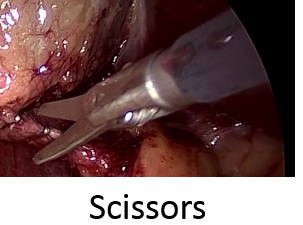} \includegraphics[width=2.5cm,height=1.9cm]{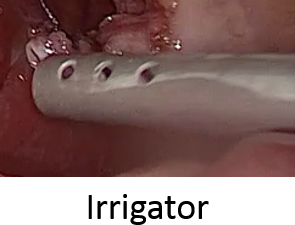}
\includegraphics[width=2.5cm,height=1.9cm]{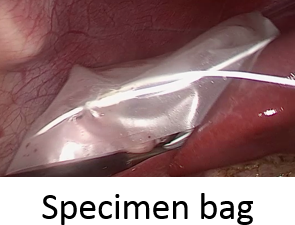}\tabularnewline
\tabularnewline
\end{tabular}
\par\end{centering}

\caption{List of the seven surgical tools used in the Cholec80 dataset. \label{fig:The-surgical-tools}}

\end{figure*}

We have constructed a large dataset, called \textit{Cholec80},
containing 80 videos of cholecystectomy surgeries performed by 13 surgeons. 
The videos are captured at 25
fps and downsampled to 1 fps for processing. The whole dataset is
labeled with the phase and tool presence annotations. The phases have been defined by a senior surgeon in our partner hospital. Since the tools are sometimes hardly visible in the images and thus difficult to be recognized visually, we define a tool as present in an image if at least half of the tool tip is visible. The tool
and the phase lists can be found in Fig. \ref{fig:The-surgical-tools}
and Tab. \ref{tab:Phase-list.}-a, respectively. 

\begin{figure}
\begin{centering}
\begin{tabular}{c}
\includegraphics[width=7cm]{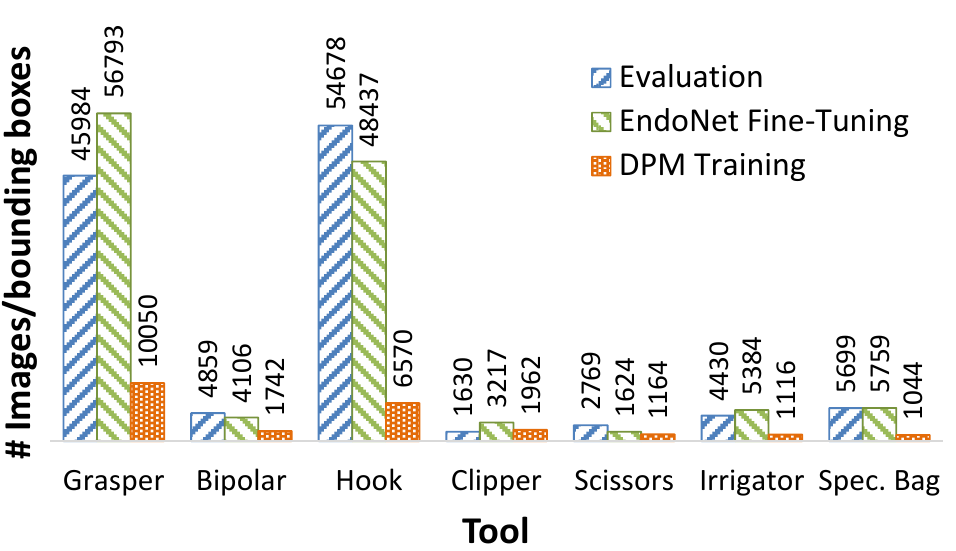}\tabularnewline
{\scriptsize (a)}\tabularnewline
\end{tabular}
\par\end{centering}

\begin{centering}
\begin{tabular}{c}
\includegraphics[width=7cm]{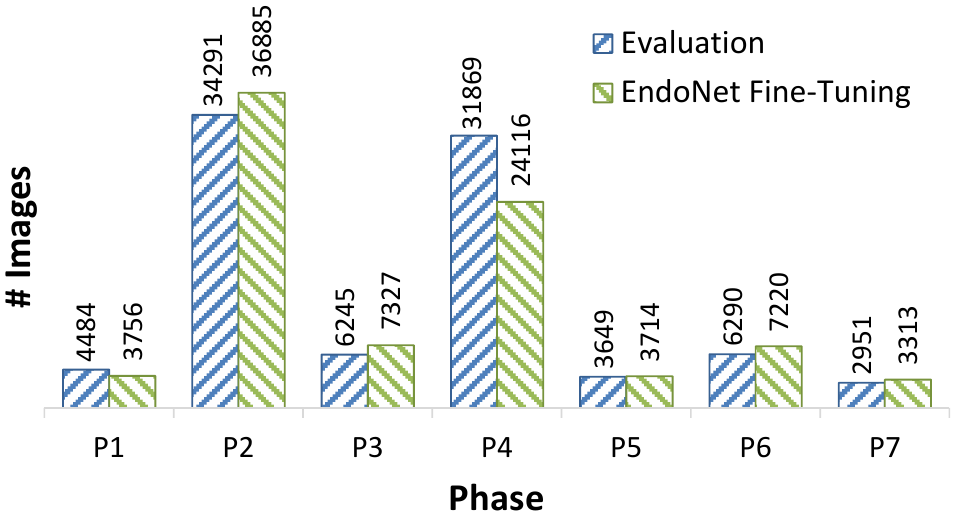}\tabularnewline
{\scriptsize (b)}\tabularnewline
\end{tabular}
\par\end{centering}

\caption{Distribution of annotations in the Cholec80 dataset for (a) tool presence
detection and (b) phase recognition tasks. \label{fig:Dataset-statistics.}}

\end{figure}

The Cholec80 dataset is split into two subsets of equal size (i.e.,
40 videos each). The first subset (i.e., the fine-tuning subset) contains \mytilde 86K annotated images.
From this subset, 10 videos have also been fully annotated with the bounding
boxes of tools. These are used to train Deformable Part Models
(DPM) \cite{DPM}.
Because the grasper and hook appear more often than other tools, their bounding boxes reach a sufficient number from the annotation of three videos. The second subset (i.e., the evaluation subset) is used to test the methods for both tool presence detection and phase recognition. The statistics of
the complete dataset can be found in Fig. \ref{fig:Dataset-statistics.}.

\begin{table}
\begin{centering}
\begin{tabular}{c}
\begin{tabular}{@{}C{1cm}L{3cm}C{1.5cm}@{}}
\toprule 
ID & Phase & Duration (s) \\ \midrule 
P1 & Preparation & 125$\pm$95\;\; \\
P2 & Calot triangle dissection & 954$\pm$538\\
P3 & Clipping and cutting & 168$\pm$152\\
P4 & Gallbladder dissection & 857$\pm$551\\
P5 & Gallbladder packaging & 98$\pm$53\\
P6 & Cleaning and coagulation & 178$\pm$166\\
P7 & Gallbladder retraction & 83$\pm$56\\
\bottomrule 
\end{tabular}\tabularnewline
{\scriptsize (a) Cholec80}\tabularnewline
\begin{tabular}{@{}C{1cm}L{3cm}C{1.5cm}@{}}
\toprule
ID & Phase & Duration (s)\tabularnewline
\midrule
P0 & Placement trocars & 180$\pm$118\tabularnewline
P12 & Preparation & 419$\pm$215\tabularnewline
P3 & Clipping and cutting & 390$\pm$194\tabularnewline
P4 & Gallbladder dissection & 563$\pm$436\tabularnewline
P5 & Retrieving gallbladder & 391$\pm$246\tabularnewline
P6 & Hemostasis & 336$\pm$62\;\;\tabularnewline
P7 & Drainage and closing & 171$\pm$128\tabularnewline
\bottomrule
\end{tabular}\tabularnewline
{\scriptsize (b) EndoVis}\tabularnewline
\end{tabular}
\par\end{centering}

\caption{List of phases in the (a) Cholec80 and (b) EndoVis datasets, including
the mean $\pm$ std of the duration of each phase in seconds. \label{tab:Phase-list.}}
\end{table}

The second dataset is a public dataset from the EndoVis workflow challenge at MICCAI 2015,
containing seven cholecystectomy videos. Similarly, these videos are captured
at 25 fps and processed at 1 fps. We only perform phase detection on this dataset, because the types and the visual appearances of the tools are different from the tools that EndoNet is designed to detect. The list of phases in the EndoVis dataset is
shown in Tab. \ref{tab:Phase-list.}-b. It can be seen that phase P3 is longer in Endovis than in Cholec80. This is due to the fact that in Cholec80, P3 is typically started when the calot triangle is clearly exposed. Yet, this is not the case in EndoVis. As a result, extra dissection steps are included in P3, leading to a longer P3 in EndoVis. 

\begin{figure}
\begin{centering}
\begin{tabular}{c}
\includegraphics[width=5cm]{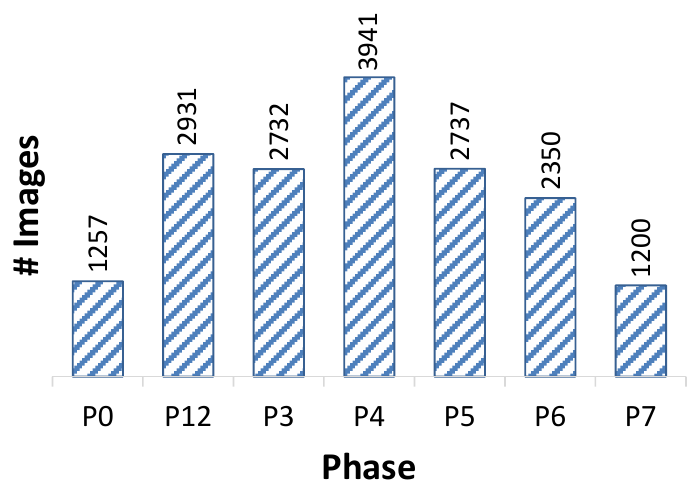}\tabularnewline
{\scriptsize (a)}\tabularnewline
\end{tabular}
\par\end{centering}

\caption{Phase distribution in the EndoVis dataset. \label{fig:EndoVis-statistics.}}

\end{figure}

The phases in EndoVis have been defined differently from the definition in Cholec80. For instance, a phase \textit{placement trocars} is defined in the EndoVis dataset, even though it should be noted that this phase is not always visible from the laparoscopic videos. Additional sources of information (e.g., external videos), which are not available in the dataset, are required to label this phase correctly. Another difference is in the definition of the \textit{preparation} phase. In the EndoVis dataset, the \textit{preparation} phase  includes the \textit{calot triangle dissection} phase (hence the ID \textit{P12} in Tab. \ref{tab:Phase-list.}-b). The other phases are defined similarly to the phases in Cholec80. The distribution of the phases in EndoVis is shown in Fig. \ref{fig:EndoVis-statistics.}. 

\subsection{Fine-Tuning, SVM and HHMM Parameters}

\begin{figure}
\begin{centering}
\includegraphics[width=8cm]{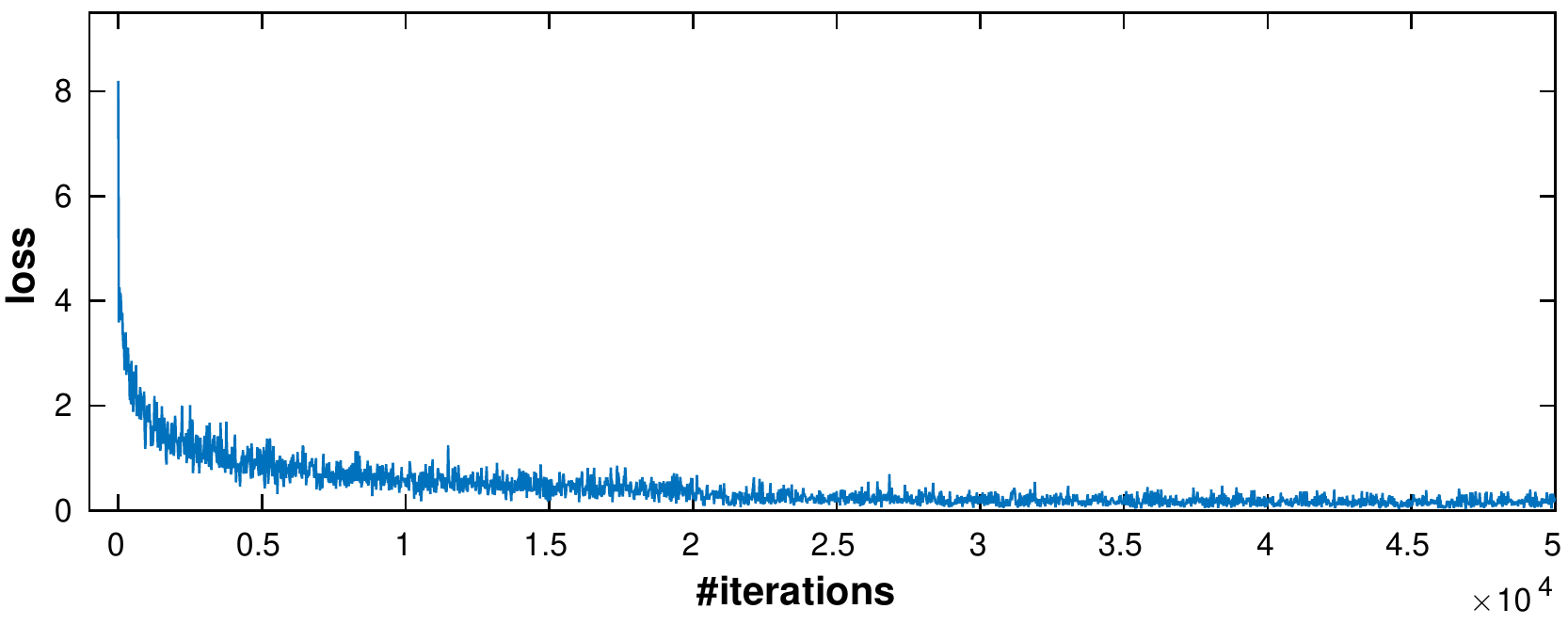}
\par\end{centering}

\caption{Evolution of the loss function during the fine-tuning process of EndoNet. \label{fig:The-evolution-of}}
\end{figure}

EndoNet is trained by fine-tuning the AlexNet network \cite{krizhevsky_nips2012} which has been pre-trained on the ImageNet dataset \cite{imagenet}. The layers that are not defined in AlexNet (i.e., $\mathtt{fc}$\_$\mathtt{tool}$ and $\mathtt{fc}$\_$\mathtt{phase}$) are initialized randomly. 
The network is fine-tuned for 50K iterations with $N_{i}=50$ images in a batch.
The learning rate is initialized at $10^{-3}$ for all layers, except
for $\mathtt{fc}$\_$\mathtt{tool}$ and $\mathtt{fc}$\_$\mathtt{phase}$,
whose learning rate is set higher at $10^{-2}$ because of their random initialization. The learning rates for all layers decrease by a factor of
$10$ for every 20K iterations. The fine-tuning process is carried
out using the Caffe framework \cite{caffe}. The evolution of the loss function $\mathcal{L}$ during the fine-tuning process
is shown in Fig. \ref{fig:The-evolution-of}. The graph shows the
convergence of the loss, indicating that the network is successfully optimized
to learn the optimal features for the phase recognition and tool presence
detection tasks.

The networks are trained using an NVIDIA GeForce Titan X graphics card. The training process takes \mytilde{80} seconds for 100 iterations, i.e., roughly 11 hours per network. The feature extraction process takes approximately 0.2 second per image. The computational time for SVM training depends on the size of the features, ranging from 0.1 to 90 seconds, while the HHMM training takes approximately 15 seconds using our MATLAB implementation.

To carry out phase recognition, all features are passed to a one-vs-all \textit{linear} SVM, except the handcrafted features,
which are passed through a histogram intersection kernel beforehand. We tried
to use non-linear kernels for other features in our preliminary experiments,
but this did not yield any improvements. 

\begin{figure}
\begin{centering}
\includegraphics[width=6cm]{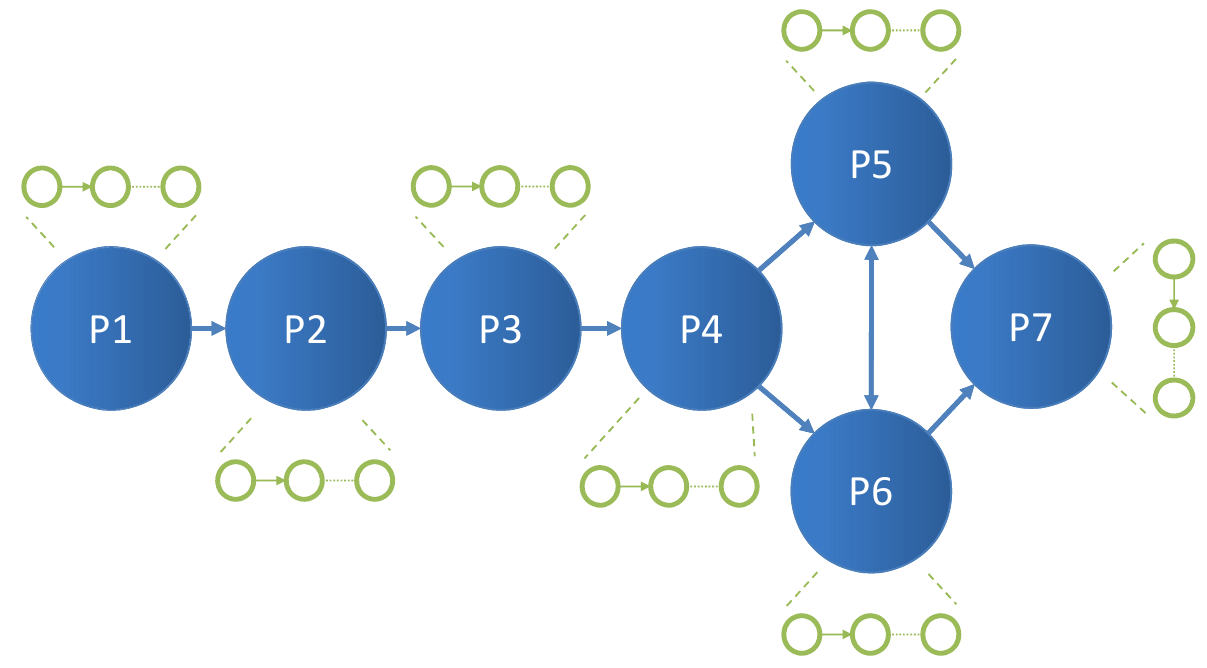}
\par\end{centering}

\caption{Graph representation of the two-level HHMM for the surgical phases defined in Cholec80. The top-level states, representing the phases defined in the dataset, are shown in blue. The transitions for top-level states show all possible phase transitions defined in the dataset. The bottom-level states are shown in green. \label{fig:HHMM} }
\end{figure}

For the HHMM, we set the number of top-level states to seven (equal to $N_{p}$), while the number of bottom-level states is data-driven (as in \cite{padoy_cvw2009}). To model the output of the SVM, we use a mixture of five Gaussians for every feature, except for the binary tool signal, where one Gaussian is used. The type of covariance is diagonal. In Fig. \ref{fig:HHMM}, the graph representation of the HHMM used to recognize the phases in Cholec80 is shown. 

\subsection{Baselines}

\begin{figure}
\begin{centering}
\begin{tabular}{ccc}
\includegraphics[width=3.5cm]{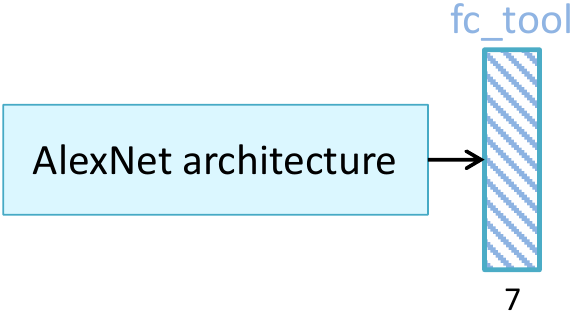} &  & \includegraphics[width=3.5cm]{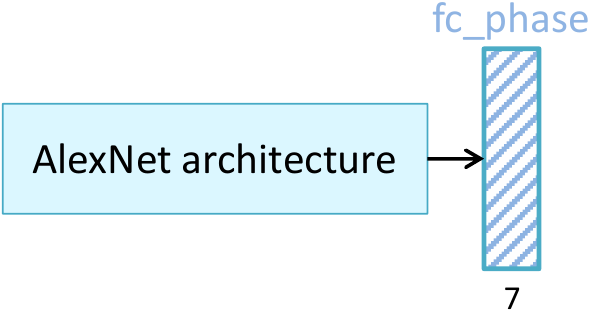}\tabularnewline
{\scriptsize (a) ToolNet} &  & {\scriptsize (b) PhaseNet}\tabularnewline
\end{tabular}
\par\end{centering}

\caption{Single-task CNN architectures for the (a) tool presence detection 
and (b) phase recognition tasks. The AlexNet architecture is the same
as the one used in EndoNet (see Fig. \ref{fig:CammaNet-Multi-architecture.}). The single-task networks are also trained via transfer learning.
\label{fig:Single-task-CNN-architectures}}
\end{figure}

For tool presence detection, we compare the results given by
EndoNet (i.e., the output of layer $\mathtt{fc}$\_$\mathtt{tool}$)
with two other methods. The first method is DPM \cite{DPM}, since
it is an ubiquitous method for object detection that is available
online. 
In the experiments, we use the default parameters, model each tool using three components and represent the images using HOG features. The second method is a network trained in a single-task manner that solely performs the tool presence detection task (ToolNet). We compare the ToolNet results with the EndoNet results in order to show that performing the fine-tuning process in a multi-task manner yields a better network than in a single-task manner. The architecture of this network can be seen in Fig. \ref{fig:Single-task-CNN-architectures}-a. 

For phase recognition, we run a 4-fold cross-validation on  the evaluation subset of Cholec80
and full cross-validation on the EndoVis dataset. Because the 
recognition pipeline contains methods trained with random initializations, the results might be 
different in each run. Thus, the displayed results are
the average of five experimental runs. Here, we compare the phase recognition results using the following features as input:
\begin{itemize}
\item binary tool information generated from the manual annotation; this is a vector
depicting the presence of the tools in an image, i.e. $\mathbf{v}_{t}\in\left\{ 0,1\right\} ^{7}$ and $\mathbf{v}_{t}\in\left\{ 0,1\right\} ^{10}$ for the Cholec80 and EndoVis datasets, respectively;
\item handcrafted visual features: bag-of-word of SIFT,
HOG, RGB and HSV histograms; these features are chosen because they have been successful in carrying out classification \cite{twinanda_ipcai2014} on laparoscopic videos;
\item the afore-mentioned handcrafted visual features + CCA, similar to the approach suggested in \cite{blum_miccai2010};
\item the output of layer $\mathtt{fc7}$ of AlexNet trained on the ImageNet
dataset (i.e., the initialization of the fine-tuning process);
\item the output of layer $\mathtt{fc7}$ from a network that is fine-tuned
to carry out phase recognition in a single-task manner, shown
in Fig. \ref{fig:Single-task-CNN-architectures}-b (PhaseNet);
\item our proposed features, i.e., the output of layer $\mathtt{fc8}$ from EndoNet.
\end{itemize}
We also include features called  EndoNet-GTbin for phase recognition on the Cholec80 dataset. 
These features consist of the output of layer $\mathtt{fc7}$ from EndoNet 
concatenated with binary tool information obtained from the ground-truth annotations.
This evaluation allows us to investigate whether the tool information automatically extracted from EndoNet, which is included in our proposed features, is sufficient for the phase recognition task.

\subsection{Evaluation}

The performance of the tool presence detection is measured by the
average precision (AP) metric. It is obtained by computing the area under the precision-recall curve. For the phase recognition task, several
evaluation metrics are used, i.e., precision, recall, and accuracy as defined in \cite{padoy_mia2012}. 
Recall and precision compute the number of correct detections divided 
by the length of the ground truth and by the length of the complete detections, respectively. 
Since they are computed for each phase, we show the averages for recall and 
precision to present summarized results. Accuracy represents the percentage of 
correct detections in the complete surgery. 

In order to show the improvements that the proposed features yield, we compute the evaluation metrics for phase recognition on the results before and after applying HHMM. To provide a deeper analysis of the results, we also present in Section \ref{sect:applications} the performance of EndoNet on two practical applications.

\section{Results}
\label{sec:Exps}

\subsection{Cholec80 Dataset}

\subsubsection{\textbf{Tool Presence Detection}} The results of the tool presence detection task are shown in Tab. \ref{tab:Tool-presence-detection}.
It can be seen 
that the networks yield significantly better results than DPM. 
It might be due to the fact that the number of images used for fine-tuning the networks is higher than the number of bounding boxes used for DPM training, but this may only partly explain this large difference. To provide a fairer comparison, we compare the performance of DPM with ToolNet and EndoNet models that are trained only with the 10 videos used to train DPM (see also Subsubsection \ref{subsect:effect-of-fine-tuning-size} for the influence of the fine-tuning subset size). As expected, the performance of the networks is lower compared to the networks trained on the full fine-tuning subset. However, the mean APs are still better than the one of DPM: 65.9 and 62.0 for ToolNet and EndoNet, respectively.  Note that, the networks are only trained using binary annotations (present vs. not-present), while DPM uses bounding boxes containing specific localization  information. Furthermore, the networks contain a much higher number of unknowns to optimize than DPM. In spite of these facts, with the same amount of training data, the networks perform the task better than DPM.

From Tab. \ref{tab:Tool-presence-detection}, it can be seen that EndoNet gives the best results for this task.
This shows that training the network in a multi-task manner does not compromise 
the EndoNet's performance in detecting the tool presence. 
For all methods, there is a decrease in performance for scissors detection. This might be due to the fact that this tool has the smallest amount of training data (see Fig. \ref{fig:Dataset-statistics.}-a), as it only appears shortly in the surgeries. In addition, it could be confused with the grasper since they share many visual similarities.
Over the seven tools and 40 complete surgeries in the evaluation subset of Cholec80, EndoNet obtains
 81\% mean AP for tool presence detection. The
success of this network
suggests that  binary annotations are sufficient to train a model for this task.
This is particularly interesting, since
tagging the images with binary information of tool presence is
much easier than providing bounding boxes. It also shows
that the networks can successfully detect tool presence without any explicit localization pre-processing steps (such as segmentation and
ROI selection). 

\begin{table}
\begin{centering}
\begin{tabular}{|c|c|c|c|}
\hline 
Tool & DPM & ToolNet & EndoNet\tabularnewline
\hline 
\hline 
Bipolar & 60.6 & 85.9 & \textbf{86.9}\tabularnewline
\hline 
Clipper & 68.4 & 79.8 & \textbf{80.1}\tabularnewline
\hline 
Grasper & 82.3 & 84.7 & \textbf{84.8}\tabularnewline
\hline 
Hook & 93.4 & 95.5 & \textbf{95.6}\tabularnewline
\hline 
Irrigator & 40.5 & 73.0 & \textbf{74.4}\tabularnewline
\hline 
Scissors & 23.4 & \textbf{60.9} & 58.6\tabularnewline
\hline 
Specimen bag & 40.0 & 86.3 & \textbf{86.8}\tabularnewline
\hline 
MEAN & 58.4 & 80.9 & \textbf{81.0}\tabularnewline
\hline 
\end{tabular}
\par\end{centering}

\caption{Average precision (AP) for all tools, computed on the 40 videos forming the evaluation dataset of Cholec80. The best AP for each tool is written in bold.
\label{tab:Tool-presence-detection}}
\end{table}

\subsubsection{\textbf{Phase Recognition}} In Tab. \ref{tab:frame-wise-recognition}-a, the results of phase recognition on Cholec80 before applying HHMM are shown. These are the results after passing the image features to the SVM. The results show that the CNNs are powerful tools to extract visual
features: despite being trained on a
completely unrelated dataset, the AlexNet features outperform the handcrafted visual 
features (without and with CCA) and the binary tool annotation. Furthermore, the
fine-tuning step significantly improves the results: the PhaseNet features yield improvements for all metrics compared to the AlexNet features. In addition to yielding the tool presence detection as a by-product, the multi-task framework
applied in EndoNet further improves the features for the phase recognition task. It is also interesting to observe that the phase recognition results using the EndoNet-GTbin features are only slightly better than the ones using the EndoNet features, with approximately 0.1\% improvement in accuracy. In other words, the tool information generated from the ground-truth does not bring more information than the EndoNet features and the visual features extracted by EndoNet alone are sufficient to carry out the phase recognition task.

In Tab. \ref{fig:Phase-recognition-results-CC}, the phase recognition
results after applying HHMM are shown. Due to the
nature of offline phase recognition, where the algorithm can see the complete video, 
the offline results are better
than the online counterparts. However, when we compare the feature performance,
the trend is consistent across the offline and online modes. By comparing the results from Tab \ref{tab:frame-wise-recognition}-a and Tab \ref{fig:Phase-recognition-results-CC}-a, we can see the improvement that the HHMM brings, which is consistent across all features.

\begin{figure}
\begin{centering}
\begin{tabular}{@{}C{4.2cm}C{4.2cm}@{}}
\includegraphics[width=4cm]{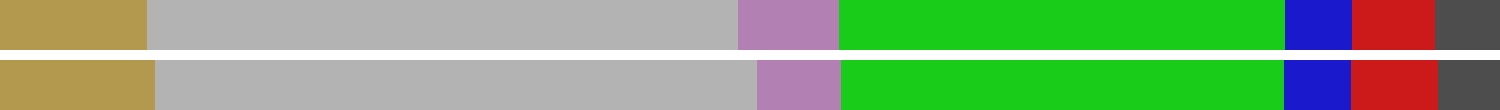} & \includegraphics[width=4cm]{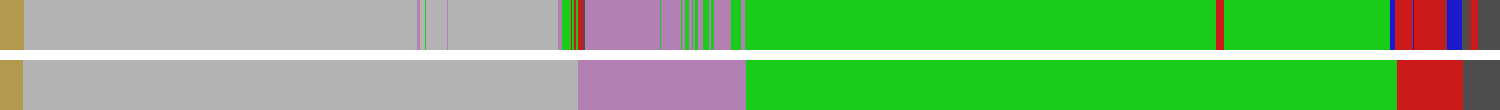}\tabularnewline
\includegraphics[width=4cm]{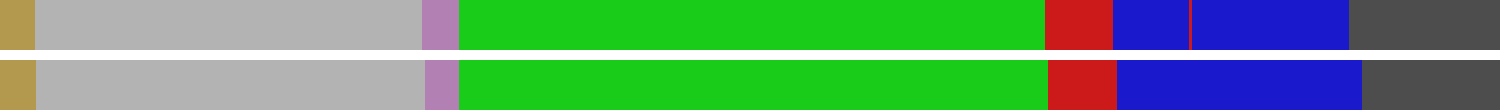} & \includegraphics[width=4cm]{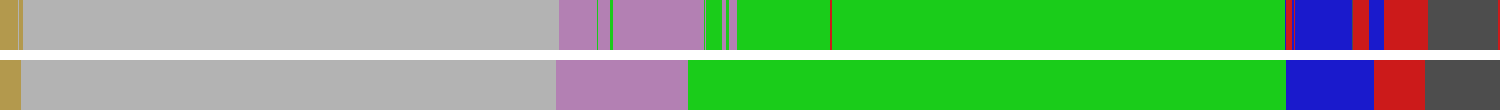}\tabularnewline
\includegraphics[width=4cm]{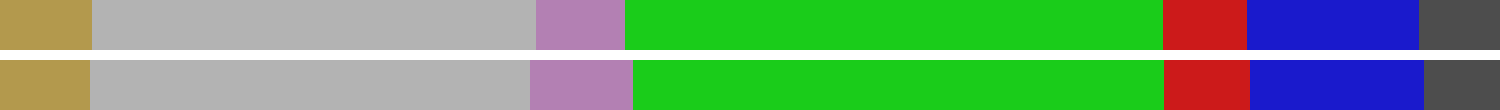} & \includegraphics[width=4cm]{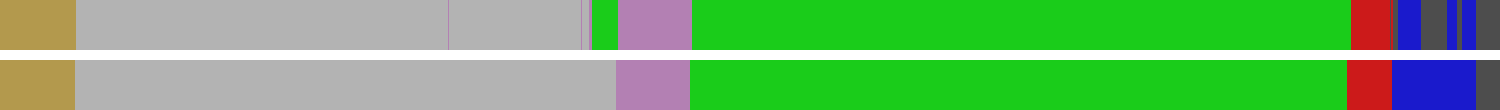}\tabularnewline
\includegraphics[width=4cm]{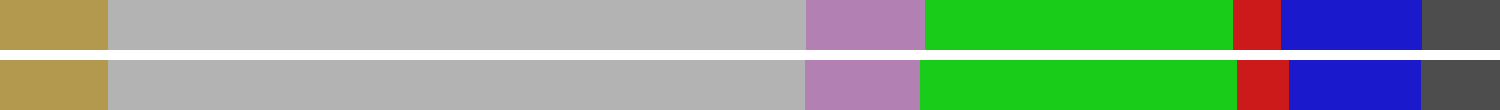} & \includegraphics[width=4cm]{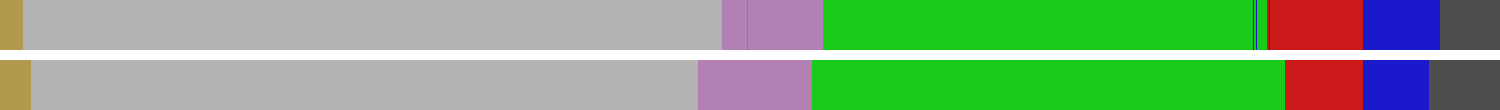}\tabularnewline
\includegraphics[width=4cm]{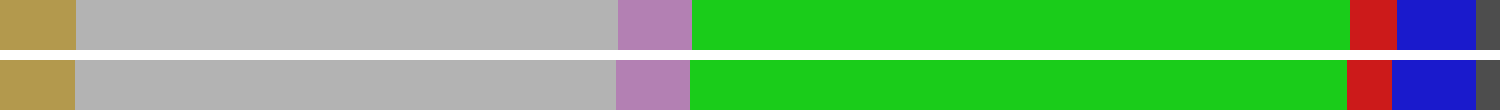} & \includegraphics[width=4cm]{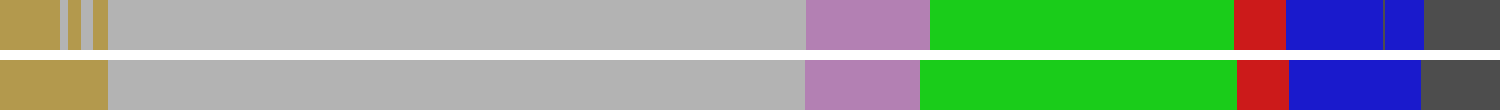}\tabularnewline
{\scriptsize (a) Top-5 offline} & {\scriptsize (b) Top-5 online}\tabularnewline
\includegraphics[width=4cm]{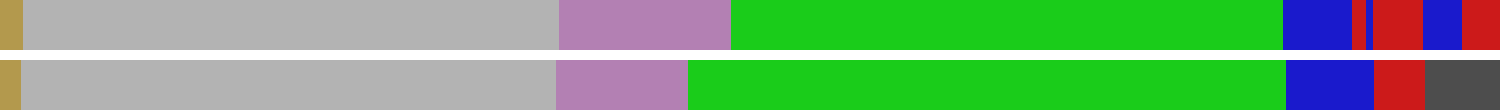} & \includegraphics[width=4cm]{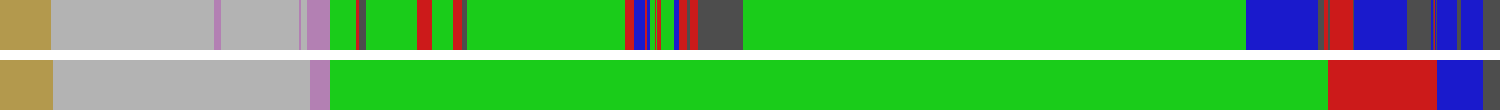}\tabularnewline
\includegraphics[width=4cm]{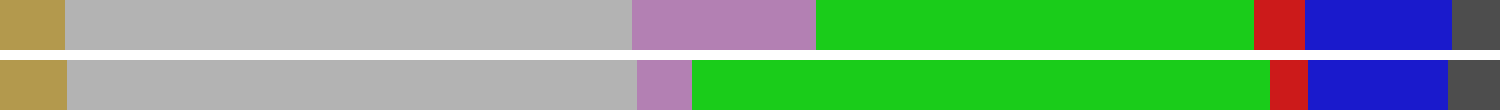} & \includegraphics[width=4cm]{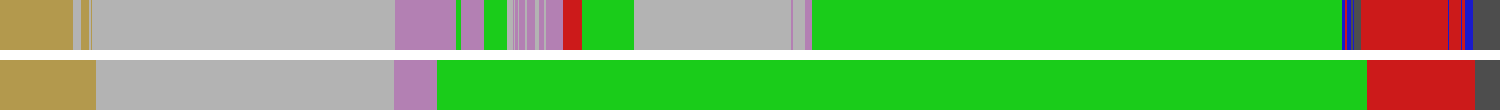}\tabularnewline
\includegraphics[width=4cm]{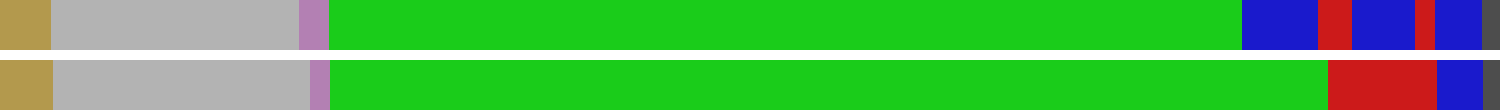} & \includegraphics[width=4cm]{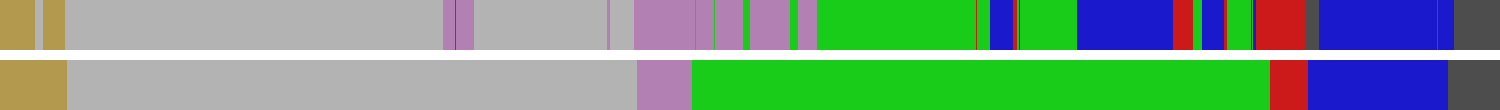}\tabularnewline
\includegraphics[width=4cm]{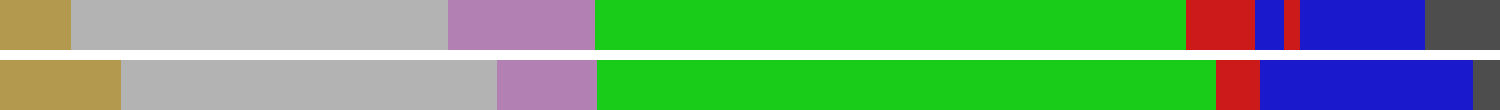} & \includegraphics[width=4cm]{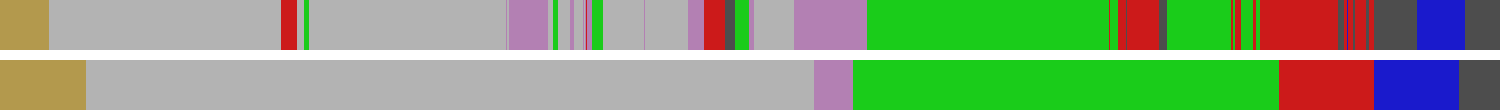}\tabularnewline
\includegraphics[width=4cm]{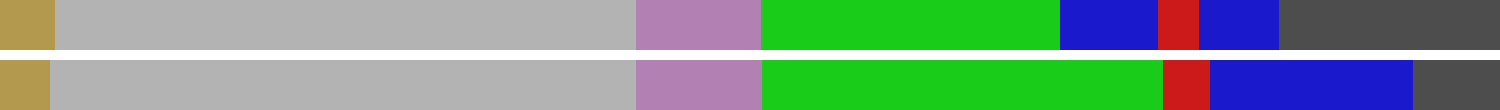} & \includegraphics[width=4cm]{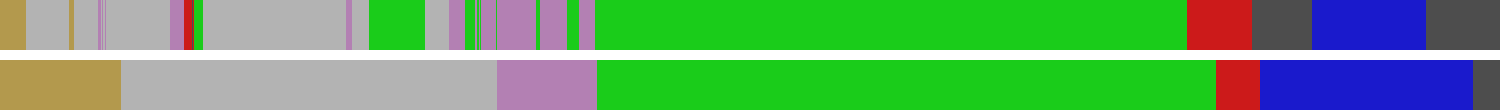}\tabularnewline
{\scriptsize (c) Bottom-5 offline} & {\scriptsize (d) Bottom-5 online}\tabularnewline
\end{tabular}

\begin{tabular}{@{}C{1cm}C{1cm}C{1cm}C{1cm}C{1cm}C{1cm}C{1cm}@{}}
\includegraphics[width=1cm,height=0.2cm]{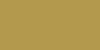} &
\includegraphics[width=1cm,height=0.2cm]{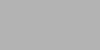} &
\includegraphics[width=1cm,height=0.2cm]{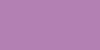} &
\includegraphics[width=1cm,height=0.2cm]{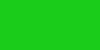} &
\includegraphics[width=1cm,height=0.2cm]{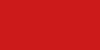} &
\includegraphics[width=1cm,height=0.2cm]{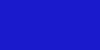} &
\includegraphics[width=1cm,height=0.2cm]{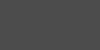} \tabularnewline
{\scriptsize P1} & {\scriptsize P2} & {\scriptsize P3} & {\scriptsize P4} & {\scriptsize P5} & {\scriptsize P6} & {\scriptsize P7} \tabularnewline
\end{tabular}
\par\end{centering}

\caption{Phase recognition results vs. ground truth on Cholec80 in a color-coded ribbon illustration. The horizontal axis of the ribbon represents the time progression in a surgery. The top ribbon is the estimated phase and the bottom ribbon is the ground truth. \label{fig:color-coded} }

\end{figure}

In Fig. \ref{fig:color-coded}, we show the top-5 and bottom-5 recognition results based on the accuracy from one (randomly chosen) experimental run in both offline and online modes. In offline mode, it can be seen that the top-5 results are very good, resulting in over 98\% accuracies. In addition, the bottom-5 results in offline mode are comparable to the ground truth. The drop of accuracy for the bottom-5 are caused by the jumps that can happen between P5 and P6, which are shown by the alternating blue and red in Fig. \ref{fig:color-coded}-c.  These jumps occur because of the non-linear transitions among these phases (see Fig. \ref{fig:HHMM}).

In online mode, one can observe more frequent jumps in the phase estimations. This is due to the nature of recognition in online mode, where future data is unavailable, so that the model is allowed to correct itself after making an estimation. Despite these jumps, the top-5 online results are still very close to the ground-truth, resulting in accuracies above 92\%.

In order to provide more comprehensive information regarding the performance of EndoNet over the whole dataset, we present the recognition results for all phases in both offline and online modes in Tab. \ref{tab:Phase-recognition-results-each-phase}. It can be seen that the EndoNet features perform very well in recognizing all the phases. A decrease in performance can be observed for the recognition of P5 and P6. 
This is likely due to the fact that the transitions between these phases are not sequential and that there is not always a clear boundary between them, especially as some images sometimes do not show any activity. This creates some ambiguity in the phase estimation process.

\begin{table*}
\begin{centering}
\begin{tabular}{cc}
\begin{tabular}{|C{2.5cm}|C{1.7cm}|C{1.7cm}|C{1.7cm}|}
\hline 
\multirow{2}{*}{Feature} & \multicolumn{3}{c|}{Cholec80}\tabularnewline
\cline{2-4} 
 & Avg. Precision & Avg. Recall & Accuracy\tabularnewline
\hline 
\hline 
Tool binary & 42.8$\pm$33.9 & 41.1$\pm$32.3 & 48.2$\pm$2.7\tabularnewline
\hline 
Handcrafted & 22.7$\pm$28.8 & 17.9$\pm$28.9 & 44.0$\pm$1.8\tabularnewline
\hline 
Handcrafted+CCA & 21.9$\pm$14.1 & 18.7$\pm$23.3 & 39.0$\pm$0.6\tabularnewline
\hline 
AlexNet & 50.4$\pm$12.0 & 44.0$\pm$22.5 & 59.2$\pm$2.4\tabularnewline
\hline 
PhaseNet & 67.0$\pm$9.3\;\; & 63.4$\pm$11.8 & 73.0$\pm$1.6\tabularnewline
\hline 
EndoNet & 70.0$\pm$8.4\;\; & 66.0$\pm$12.0 & 75.2$\pm$0.9\tabularnewline
\hline 
EndoNet+GTBin & 70.1$\pm$9.1\;\; & 66.7$\pm$11.1 & 75.3$\pm$1.1\tabularnewline
\hline 
\end{tabular} & %
\begin{tabular}{|C{1.7cm}|C{1.7cm}|C{1.7cm}|}
\hline 
\multicolumn{3}{|c|}{EndoVis}\tabularnewline
\hline 
Avg. Precision & Avg. Recall & Accuracy\tabularnewline
\hline 
\hline 
44.3$\pm$32.5 & 48.5$\pm$39.3 & 49.0$\pm$9.7\tabularnewline
\hline 
35.7$\pm$6.6\;\; & 33.2$\pm$10.5 & 36.1$\pm$2.6\tabularnewline
\hline 
31.1$\pm$4.6\;\; & 31.6$\pm$22.6 & 32.6$\pm$5.3\tabularnewline
\hline 
60.2$\pm$8.0\;\; & 57.8$\pm$9.3\;\; & 56.9$\pm$4.1\tabularnewline
\hline 
63.5$\pm$5.7\;\; & 63.2$\pm$9.3\;\; & 62.6$\pm$4.9\tabularnewline
\hline 
64.8$\pm$7.3\;\; & 64.3$\pm$11.8 & 65.9$\pm$4.7\tabularnewline
\hline 
\multicolumn{1}{c}{} & \multicolumn{1}{c}{} & \multicolumn{1}{c}{}\tabularnewline
\end{tabular}\tabularnewline
\end{tabular} 
\par\end{centering}

\caption{Phase recognition results before applying the HHMM (mean $\pm$ std) on Cholec80 and EndoVis. \label{tab:frame-wise-recognition}}

\end{table*} 

\begin{table*}
\begin{centering}
\begin{tabular}{c}
\begin{tabular}{@{}|C{2.5cm}|C{1.7cm}|C{1.7cm}|C{1.7cm}|@{}}
\hline 
\multirow{2}{*}{Feature} & \multicolumn{3}{c|}{Overall-Offline (\%)}\tabularnewline
\cline{2-4} 
 &  Avg. Precision & Avg. Recall & Accuracy\tabularnewline
\hline 
\hline 
Binary tool & 68.4$\pm$24.1 & 75.7$\pm$13.6 & 69.2$\pm$8.0\tabularnewline
\hline 
Handcrafted  & 40.3$\pm$20.4 & 40.0$\pm$17.8 & 36.7$\pm$7.8\tabularnewline
\hline 
Handcrafted+CCA  & 54.6$\pm$23.8 & 57.2$\pm$21.2 & 61.3$\pm$8.3\tabularnewline
\hline 
AlexNet  & 70.9$\pm$12.0 & 73.3$\pm$16.7 & 76.2$\pm$6.3\tabularnewline
\hline 
PhaseNet  & 82.5$\pm$9.8\;\; & 86.6$\pm$4.5\;\; & 89.1$\pm$5.4\tabularnewline
\hline 
EndoNet & \textit{84.8}$\pm$\textit{9.1}\;\; & \textit{88.3}$\pm$\textit{5.5}\;\; & \textit{92.0}$\pm$\textit{1.4}\tabularnewline
\hline
EndoNet-GTbin & \textbf{85.7}$\pm$\textbf{9.1}\;\; & \textbf{89.1}$\pm$\textbf{5.0}\;\; & \textbf{92.2}$\pm$\textbf{3.5} \tabularnewline 
\hline
\end{tabular} %
\begin{tabular}{@{}|C{1.7cm}|C{1.7cm}|C{1.7cm}|@{}}
\hline 
\multicolumn{3}{|c|}{Overall-Online (\%)}\tabularnewline
\hline 
Avg. Precision & Avg. Recall & Accuracy\tabularnewline
\hline 
\hline 
 54.5$\pm$32.3 & 60.2$\pm$23.8 & 47.5$\pm$2.6\tabularnewline
\hline 
 31.7$\pm$20.2 & 38.4$\pm$19.2 & 32.6$\pm$6.4\tabularnewline
\hline 
 39.4$\pm$31.0 & 41.5$\pm$21.6 & 38.2$\pm$5.1\tabularnewline
\hline 
 60.3$\pm$21.2 & 65.9$\pm$16.0 & 67.2$\pm$5.3\tabularnewline
\hline 
 71.3$\pm$15.6 & 76.6$\pm$16.6 & 78.8$\pm$4.7\tabularnewline
\hline 
 \textit{73.7}$\pm$\textit{16.1} & \textit{79.6}$\pm$\textit{7.9}\;\; & \textit{81.7}$\pm$\textit{4.2}\tabularnewline
 \hline
 \textbf{75.1}$\pm$\textbf{15.6}  & \textbf{80.0}$\pm$\textbf{6.7} &  \textbf{81.9}$\pm$\textbf{4.4} \tabularnewline
\hline 
\end{tabular}\tabularnewline
{\scriptsize (a) Cholec80}\tabularnewline
\begin{tabular}{@{}|C{2.5cm}|C{1.7cm}|C{1.7cm}|C{1.7cm}|@{}}
\hline 
\multirow{2}{*}{Feature} & \multicolumn{3}{c|}{Overall-Offline (\%)}\tabularnewline
\cline{2-4} 
 &  Avg. Precision & Avg. Recall & Accuracy\tabularnewline
\hline 
\hline 
Binary tool & 81.4$\pm$16.1 & 79.5$\pm$12.3 & 73.0$\pm$21.5\tabularnewline
\hline 
Handcrafted & 49.7$\pm$15.6 & 33.2$\pm$21.5 & 46.5$\pm$24.6\tabularnewline
\hline 
Handcrafted+CCA & 66.1$\pm$22.3 & 64.7$\pm$22.1 & 61.1$\pm$17.3\tabularnewline
\hline 
AlexNet & 85.7$\pm$13.2 & 80.8$\pm$10.4 & 79.5$\pm$11.0\tabularnewline
\hline 
PhaseNet & 86.8$\pm$14.2 & 83.1$\pm$10.6 & 79.7$\pm$12.2\tabularnewline
\hline 
EndoNet & \textit{\textbf{91.0}}$\pm$\textit{\textbf{7.7}}\;\; & \textit{\textbf{87.4}}$\pm$\textit{\textbf{10.3}} & \textit{\textbf{86.0}}$\pm$\textit{\textbf{6.3}}\;\;\tabularnewline
\hline 
\end{tabular} %
\begin{tabular}{@{}|C{1.7cm}|C{1.7cm}|C{1.7cm}|@{}}
\hline 
 \multicolumn{3}{|c|}{Overall-Online (\%)}\tabularnewline
\hline 
 Avg. Precision & Avg. Recall & Accuracy\tabularnewline
\hline 
\hline 
 80.3$\pm$18.1 & 77.5$\pm$18.8 & 69.8$\pm$21.7\tabularnewline
\hline 
 46.6$\pm$16.2 & 48.0$\pm$18.5 & 43.4$\pm$21.6\tabularnewline
\hline 
 52.3$\pm$22.2 & 49.4$\pm$21.5 & 44.0$\pm$22.3\tabularnewline
\hline 
 78.4$\pm$14.1 & 73.9$\pm$11.4 & 70.6$\pm$12.3\tabularnewline
\hline 
 79.1$\pm$15.0 & 75.7$\pm$15.3 & 71.0$\pm$9.2\;\;\tabularnewline
\hline 
\textit{\textbf{83.0}}$\pm$\textit{\textbf{12.5}} & \textit{\textbf{79.2}}$\pm$\textit{\textbf{17.5}} & \textit{\textbf{76.3}}$\pm$\textit{\textbf{5.1}}\;\;\tabularnewline
\hline 
\end{tabular}\tabularnewline
{\scriptsize (b) EndoVis}\tabularnewline
\end{tabular} 
\par\end{centering}

\caption{Phase recognition results after applying the HHMM (mean $\pm$ std) on: (a) Cholec80
and (b) EndoVis. The best result for each evaluation metric
is written in bold. The results from our proposed features (EndoNet) are written 
in italic. \label{fig:Phase-recognition-results-CC}}
\end{table*}

\begin{table*}
\begin{centering}
\begin{tabular}{c}
\begin{tabular}{|c|c|c|c|c|c|c|c|c|}
\hline 
\multirow{1}{*}{Feature} & Metric & \multicolumn{1}{c|}{P1} & \multicolumn{1}{c|}{P2} & \multicolumn{1}{c|}{P3} & \multicolumn{1}{c|}{P4} & \multicolumn{1}{c|}{P5} & \multicolumn{1}{c|}{P6} & \multicolumn{1}{c|}{P7}\tabularnewline
\hline 
\multirow{2}{*}{EndoNet - offline} & Prec. & 83.5$\pm$9.6 & 97.1$\pm$2.0 & 81.0\textbf{$\pm$}7.7\;\; & 97.3$\pm$2.1 & 73.1$\pm$8.0\;\; & 79.7$\pm$10.4 & 81.9$\pm$11.8\tabularnewline
\cline{2-9} 
 & Rec. & 90.9$\pm$5.7 & 80.8$\pm$4.3 & 88.1\textbf{$\pm$}7.4\;\; & 94.7$\pm$1.0 & 83.7$\pm$5.6\;\; & 79.6$\pm$8.8\;\; & 86.7$\pm$11.8\tabularnewline
\hline 
\multirow{2}{*}{EndoNet - online} & Prec. & 90.0$\pm$5.6 & 96.4$\pm$2.0 & 69.8\textbf{$\pm$}10.7 & 82.8$\pm$6.2 & 55.5$\pm$11.9 & 63.9$\pm$10.5 & 57.5$\pm$11.0\tabularnewline
\cline{2-9} 
 & Rec. & 85.5$\pm$3.9 & 81.1$\pm$8.9 & 71.2\textbf{$\pm$}9.7\;\; & 86.5$\pm$4.3 & 75.5$\pm$3.8\;\; & 68.7$\pm$9.1\;\; & 88.9$\pm$7.5\;\;\tabularnewline
\hline 
\end{tabular}\tabularnewline
\end{tabular}
\par\end{centering}

\caption{Precision and recall of phase recognition for each phase on Cholec80
using the EndoNet features. \label{tab:Phase-recognition-results-each-phase}}

\end{table*}

\subsubsection{\textbf{Effects of Fine-Tuning Subset Size} \label{subsect:effect-of-fine-tuning-size}}

\begin{figure}
\begin{centering}
\begin{tabular}{c}
\includegraphics[width=7cm]{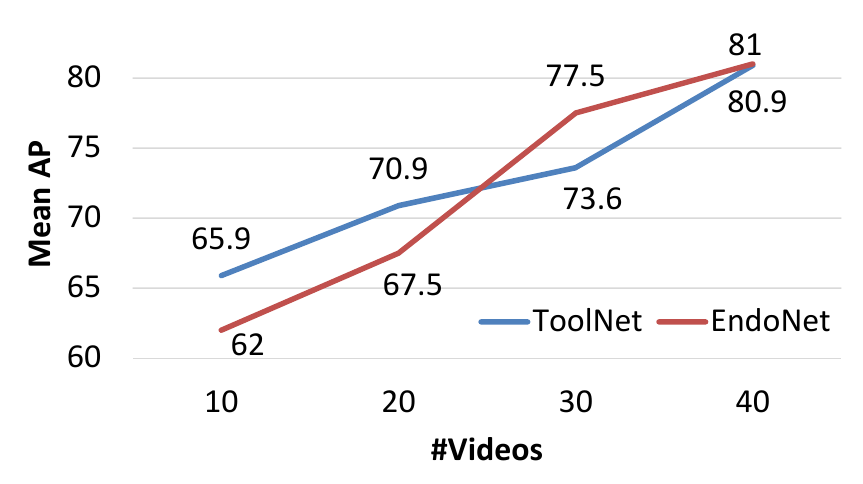}\tabularnewline
{\footnotesize (a) Tool presence detection}\tabularnewline
\includegraphics[width=7cm]{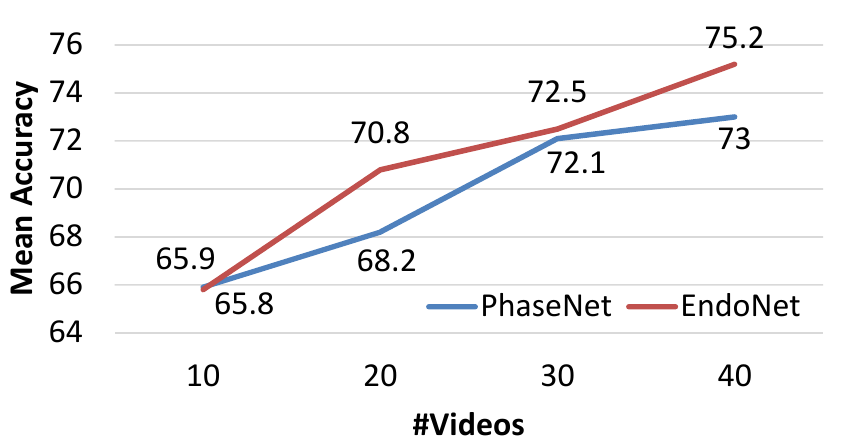}\tabularnewline
{\footnotesize (b) Phase recognition}\tabularnewline
\end{tabular}
\par\end{centering}

\caption{Evolution of network performance on Cholec80 with respect to the number of videos in the
fine-tuning subset. \label{fig:evolution-of-accuracy}}
\end{figure}

In order to show the importance of the amount of training data for the fine-tuning process, we fine-tune our networks using fine-tuning subsets with gradually increasing size: 10, 20, 30, and ultimately 40 videos. We perform both tool presence detection and phase recognition tasks on the evaluation subset of Cholec80 using the trained networks. The results are shown in Fig. \ref{fig:evolution-of-accuracy}. As expected, the performance of the networks increase proportionally to the amount of data in the fine-tuning subset. It can also be seen that EndoNet performs better than the single-task networks (i.e., PhaseNet and ToolNet), except for the tool presence detection task where fewer videos are used to train the networks. This indicates that EndoNet takes more advantage of the big dataset compared to ToolNet.

\subsection{EndoVis Dataset}

Similar results for phase recognition are obtained from the EndoVis
dataset, as shown in Tab. \ref{tab:frame-wise-recognition} and \ref{fig:Phase-recognition-results-CC}-b. It can be observed that the improvements obtained by
PhaseNet and EndoNet on EndoVis are not as high as the result improvements on Cholec80, 
which is expected since these networks are fine-tuned using the videos from Cholec80. 
In spite of this fact, the results on the EndoVis dataset also show that the EndoNet features
improve the phase recognition results significantly. It indicates that 
the multi-task learning results in a better network than the single-task 
counterpart. The fact that the features from EndoNet yield the best results for 
all cases also shows that EndoNet is generalizable to other datasets. 

One should note that we use the output
of layer $\mathtt{fc8}$ from EndoNet as the image feature, which
includes  confidence values for tool presence. Because the tools used
in EndoVis dataset are not the same tools as the ones in the Cholec80
dataset (which is used to train EndoNet), these confidence values can simply
be regarded as 7 additional scalar features appended to the feature vector. The results show that these values help to construct more discriminative features. 

\section{Medical Applications \label{sect:applications}}

Here, we demonstrate the applicability of EndoNet for 
practical CAI applications. We present the results from the same experimental 
run that is used to generate Fig. \ref{fig:color-coded}. \textbf{First}, 
to show the feasibility of using EndoNet as the basis for automatic 
surgical video indexing, we show the error of the phase estimation 
in seconds to indicate how precise the phase boundary estimations from EndoNet are. 
\textbf{Second}, we investigate further how accurately EndoNet detects the presence of two tools: clipper and bipolar. These tools are particularly interesting because: (1) 
the appearance of the clipper typically marks the beginning of the 
\textit{clipping and cutting} phase, which is the most delicate phase in the procedure, 
and (2) the bipolar tool is generally used to stop haemorrhaging, which could lead to 
possible upcoming complications.

\subsection{Automatic Surgical Video Database Indexing}
For automatic video indexing, the task corresponds to carrying out phase recognition in 
offline mode. 
From the results shown 
in Fig. \ref{fig:color-coded}-a,c, one can already roughly interpret how accurate the phase 
recognition results are. To give a more intuitive evaluation, 
we present the number of phase boundaries that are 
detected within defined temporal tolerance values in Tab. \ref{tab:offline-error-second}. 
We can see that EndoNet generally performs very well for all the phases, 
resulting in 89\% of the phase boundaries being detected within 30 seconds.
It can also be seen that only 6\% of the phase boundaries are detected with an error
over 2 minutes. It is also important to note that this error is computed with respect to the 
strict phase boundaries defined in the annotation. In practice, 
these boundaries are not as harsh or visually obvious. Thus, this 
error is  acceptable in most cases. In other words, 
it indicates that the results from EndoNet do not require a lot of corrections, 
which will make surgical video indexing a lot faster and easier. 

\begin{table}
\begin{centering}
\begin{tabular}{|c|c|c|c|c|c|c|c|}
\hline 
\multirow{2}{*}{Tolerance (s)} & \multicolumn{7}{c|}{Phase}\tabularnewline
\cline{2-8} 
 & P1 & P2 & P3 & P4 & P5 & P6 & P7\tabularnewline
\hline 
\hline 
$<$30 & 40 & 34 & 34 & 34 & 40 & 30 & 33\tabularnewline
\hline 
30-59 & 0 & 0 & 0 & 0 & 0 & 0 & 0\tabularnewline
\hline 
60-89 & 0 & 4 & 1 & 0 & 0 & 1 & 3\tabularnewline
\hline 
90-119 & 0 & 0 & 1 & 2 & 0 & 0 & 2\tabularnewline
\hline 
$\geq$120 & 0 & 2 & 4 & 4 & 0 & 4 & 2\tabularnewline
\hline 
\textbf{TOTAL} & 40 & 40 & 40 & 40 & 40 & 35 & 40\tabularnewline
\hline 
\end{tabular}
\par\end{centering}

\caption{Number of phases that are correctly identified in offline mode within the defined tolerance values in the 40 evaluation videos of Cholec80. The number of P6 occurrences is not 40 since not all surgeries go through the cleaning and coagulation phase. \label{tab:offline-error-second}}
\end{table}

%

\subsection{Bipolar and Clipper Detection}

\begin{table}
\begin{centering}
\begin{tabular}{|c|c|c|}
\hline 
Tolerance (s) & Bipolar & Clipper\tabularnewline
\hline 
\hline 
$<$5 & 114 & 49\tabularnewline
\hline 
6-29 & 9 & 10\tabularnewline
\hline 
30-59 & 1 & 0\tabularnewline
\hline 
$\geq$60 & 0 & 1\tabularnewline
\hline 
Missed & 0 & 1\tabularnewline
\hline 
False positives & 3.8\% & 8.3\%\tabularnewline
\hline 
\end{tabular}
\par\end{centering}

\caption{Appearance block detection results for bipolar and clipper, including the number of correctly classified blocks and missed blocks, and the false positive rate of the detection. \label{tab:block-detection} }
\end{table}

In addition to showing the AP for detection of both tools in Tab. \ref{tab:Tool-presence-detection}, 
we present a more intuitive metric to measure the reliability of EndoNet 
for the bipolar and clipper presence detections. We define a \textit{tool block} as a set of consecutive 
frames in which a certain tool is present.  Since the tools might not always 
be visible in an image even though they are currently being used, we merge the blocks (of the same tool) in the ground-truth data that have a gap that is less than 15 seconds. 
Then, we define a tool block as identified if EndoNet can detect the tool in 
at least one of the frames inside the block. To show the performance of 
EndoNet in terms of temporal precision, we also present the time difference between the 
first frame of the tool block and the first frame of the detection. 
In this experiment, we determine the tool presence by taking a confidence threshold that 
gives a high precision for each tool, so that the system can obtain the 
minimal amount of false positives and retain the sensitivity in correctly detecting 
the tool blocks. Since the false positive rate is measured using the tool block 
definition, we also close the gaps between the tool presence detections 
that are less than 15 seconds.

We show the block detection results in Tab. \ref{tab:block-detection}. 
It can be seen that all the bipolar blocks are detected very 
well by EndoNet. Over 90\% of the blocks are detected under 5 seconds. EndoNet also yields a very low false positive rate (i.e., 3.8\%) for the bipolar. 
This excellent performance is obtained thanks to the distinctive visual appearance that the bipolar has (e.g., the blue shaft). For the clipper, it can 
be seen that the false positive rate is higher than for the bipolar. This could 
be due to the fact that it has the second lowest amount of annotations 
in the dataset, because, similarly to the scissors, the clipper only appears shortly in the surgeries. However, EndoNet still performs very well for clipper detection, showing that 80\% and 97\% of the blocks are detected under 5 and 30 seconds, respectively.

\section{Discussion and Conclusions}

In this paper, we address the problem of phase recognition in laparoscopic surgeries and propose a novel method to learn visual features directly from raw images.
This method is based on a convolutional neural network (CNN) architecture, called EndoNet, which is designed to perform two tasks simultaneously: tool presence detection and phase recognition. 
We show through extensive experiments that this architecture yields visual features that outperform both previously used features and the features obtained from architectures designed for a single task.
Interestingly, the EndoNet visual features also perform significantly better in the phase recognition task than binary tool signals indicating which tools can be seen in the image, even though these signals are obtained from ground truth annotations. These results therefore suggest that the images contain additional characteristics useful for recognition in addition to simple tool presence information and that these characteristics are successfully retrieved by EndoNet.
Additionally, we have shown that EndoNet also performs well on another smaller dataset, namely EndoVis, and is therefore generalizable.

To train and evaluate EndoNet, we constructed a large dataset containing 80 videos of cholecystectomy procedures performed by 13 surgeons. Even though the cholecystectomy procedure is a common focus for surgical workflow analysis, to the best of our knowledge, the cholecystectomy datasets used in previous work are limited to less than 20 surgeries. This is therefore the first large-scale study performed for these recognition tasks. This is also the first extensive comparison of the features that can be used to perform phase recognition on laparoscopic surgeries\footnote{Since no significant database is currently available to compare the approaches, to encourage open research in this direction, we will make the complete annotated video dataset as well as the trained CNN architectures available to the community upon publication of this work.}. Furthermore, it is shown by the \textit{std} of the phase durations in Tab-\ref{tab:Phase-list.}-a that the dataset in itself contains a high variability. The state-of-the-art results from EndoNet indicates that our proposed method can cope with such complexity. 

The results of varying sizes of the fine-tuning subset suggest that taking more videos from Cholec80 to fine-tune the networks will lead to better performance. However, it should be noted that the videos in Cholec80 come from one hospital, thus the complexity of the data is limited to the variability of procedure executions by surgeons from the same institution. Training a CNN network with such a dataset can lead to over-fitting and subsequently reduce the generalizability of the network. To obtain more generalizable networks, videos from other medical institutions should be included to ensure a higher variability in the dataset. The success of EndoNet in carrying out the tool presence detection and phase recognition tasks should be considered as a call for action in the community to open their data to accelerate the development of generalizable solutions for these tasks. 

We have shown the applicability of EndoNet for two different applications. These applications focus on video database management, which is one of the demands from our clinical partners. In future work, other related applications should be addressed, such as context-aware assistance during live surgeries. It will also be interesting to explore whether the features generated by EndoNet can be used to perform other tasks in laparoscopic videos, such as the estimation of the completion time of the procedure \cite{padoy_mia2012}, the classification of surgical videos \cite{twinanda_ipcai2014}, and the recognition of the anatomy.

Despite yielding state-of-the-art results, the presented phase recognition pipeline still has some limitations. For example, the phase recognition still relies on the HHMM, which is required to enforce the temporal constraints in the phase estimation. Thus, the features learnt by EndoNet do not include any temporal information present in the videos. In addition, since the HHMM is trained separately from the EndoNet fine-tuning process, the EndoNet features are not optimized on the entire phase recognition task. With additional training data, these limitations could be solved by using long short term memory (LSTM) architectures. Such an approach will form part of future efforts to improve phase recognition.

\section*{Acknowledgements}

This work was supported by French state funds managed by the ANR within
the Investissements d'Avenir program under references ANR-11-LABX-0004
(Labex CAMI), ANR-10-IDEX-0002-02 (IdEx Unistra) and ANR-10-IAHU-02
(IHU Strasbourg). The authors would like to thank the IRCAD audio-visual
team for their help in generating the dataset. The authors would also
like to acknowledge the support of NVIDIA with the donation of the GPU used in this research.

\bibliographystyle{unsrt}
\bibliography{CammaNet}

\end{document}